# DETECTING INTENTIONAL LEXICAL AMBIGUITY IN ENGLISH PUNS

**Mikhalkova E. V.** (e.v.mikhalkova@utmn.ru),
**Karyakin Yu. E.** (y.e.karyakin@utmn.ru)

Tyumen State University, Tyumen, Russia

The article describes a model of automatic analysis of puns, where a word is intentionally used in two meanings at the same time (the target word). We employ Roget's Thesaurus to discover two groups of words, which, in a pun, form around two abstract bits of meaning (semes). They become a semantic vector, based on which an SVM classifier learns to recognize puns, reaching a score 0.73 for F-measure. We apply several rule-based methods to locate intentionally ambiguous (target) words, based on structural and semantic criteria. It appears that the structural criterion is more effective, although it possibly characterizes only the tested dataset. The results we get correlate with the results of other teams at SemEval-2017 competition (Task 7 Detection and Interpretation of English Puns), considering effects of using supervised learning models and word statistics.

# РАСПОЗНАВАНИЕ НАМЕРЕННОЙ ЛЕКСИЧЕСКОЙ НЕОДНОЗНАЧНОСТИ В АНГЛИЙСКИХ КАЛАМБУРАХ

**Михалькова Е. В.** (e.v.mikhalkova@utmn.ru),
**Карякин Ю. Е.** (y.e.karyakin@utmn.ru)

ФГАОУ ВО «Тюменский государственный университет», Тюмень, Россия

## 1. Concerning puns

Computational humor is a branch of computational linguistics, which developed fast in the 1990s. Its two main goals are interpretation and generation of all kinds of humor.[1] Recently we noticed a new rise of attention to this research area, especially concerning analysis of short genres like tweets [Davidov et al. 2010; Reyes et al. 2013; Castro et al. 2016]. Furthermore, a number of tasks at SemEval-2017 (an annual event, organized by the Association for Computational Linguistics) was about analyzing short funny utterances, like humorous tweets (Task 6: #HashtagWars: Learning a Sense of Humor) and puns (Task 7: Detection and Interpretation of English Puns). The following article is an extended review of the algorithm that we used for pun recognition in SemEval, Task 7.

In [Miller et al. 2015], Tristan Miller and Iryna Gurevych give a comprehensive account of what has already been done in automatic recognition of puns. They note that the study of puns mainly focused around phonological and syntactic, rather than semantic interpretation. At present, the problem of intentional lexical ambiguity is viewed more as a WSD-task, solving which is not only helpful in detecting humor, but can also provide new algorithms of sense evaluation for other NLP-systems.

The following terminology is basic in our research of puns. **A pun** is a) a short humorous genre, where a word or phrase is used intentionally in two meanings, b) a means of expression, the essence of which is to use a word or phrase so that in the given context the word or phrase can be understood in two meanings simultaneously. **A target word** is a word, used in a pun in two meanings. **A homographic pun** is a pun that "exploits distinct meanings of the same written word" [Miller et al. 2015] (these can be meanings of a polysemantic word, or homonyms, including homonymic word forms). **A heterographic pun** is a pun, in which the target word resembles another word or phrase in spelling; we will call the latter **the second target word**. More data on classification of puns and their elaborated examples can be found in [Hempelmann 2004].

(1) *I used to be a banker, but I lost interest.*

   Ex. 1 (the Banker joke) is a homographic pun; "interest" is the target word.

(2) *When the church bought gas for their annual barbecue, proceeds went from the sacred to the propane.*

   Ex. 2 (the Church joke) is a heterographic pun; "propane" is the target word, "profane" is the second target word.

Our model of automatic pun detection is based on the following premise: in a pun, there are two groups of words and their meanings that indicate the two meanings, in which the target word or phrase is used. These groups overlap, i.e. contain the same words, used in different meanings.

In Ex. 1, words and collocations "banker", "lost interest" point at the professional status of the narrator and his/her career failure. At the same time, "used to", "lost interest" tell a story of losing emotional attachment to the profession: the narrator lost curiosity. We propose an algorithm of homographic pun recognition that discovers these two groups of words and collocations, based on common semes[2], which words in these groups share. When the groups are found, in homographic puns, the next step is to state where these groups overlap, and choose which word is the target word. In case of heterographic puns, the algorithm looks for the word or phrase, which is used in one group and *not* used in the other. The last step in the analysis of heterographic puns is to calculate the second target word[3].

---

[1] In [Mikhalkova 2010] we gave a brief account of main trends in computational humor up to 2010.

## 2. Mining semantic fields

We will call a semantic field a group of words and collocations[4] that share a common seme. We hold by the opinion that the following reciprocal dependency between a word and a seme is true: in a bunch of words, the more abstract a seme is, the more words share it, and vice versa the more there are words that share a seme, the more abstract the seme is. This type of relations between lexical items can be found in taxonomies, like WordNet [Fellbaum 1998] and Roget's Thesaurus [Roget 2006] (further referred to as Thesaurus). Applying such dictionaries to get the common groups of words in a pun is, therefore, the task of finding two most general hypernyms in WordNet, or two relevant Classes among the six Classes in Thesaurus. We chose Thesaurus, as its structure is not more than five levels deep, Classes labels are not lemmas themselves, but arbitrary names (we used numbers instead), and it allows parsing on a certain level and insert corrections. After some experimentation, instead of Classes, we chose to search for relevant Sections, which are 34 subdivisions[5] of the six Classes.

(3) *I wasn't originally going to get a brain transplant, but then I changed my mind.*

Beside its structure, Thesaurus contains many collocations; these are not only multi-word units, but also aphorisms, proverbs, etc. The collocations have their own position in Thesaurus, different from the words, which compose them. Preliminary research showed the importance of collocations, in which target words appear. Sometimes the whole pun stands on rethinking a stable union of words, like in Ex. 3, "to change one's mind" becomes "to change one's brain". Therefore, when the semantic fields in a pun are discovered, it is sometimes crucial that the algorithm also analyzes collocations. In the current implementation, our program extracts collocations, based on their morphological composition. The following patterns are used: (verb+particle), (verb+(determiner/pronoun)[6]+noun+((conjunction/preposition)+noun)), (verb+adverb), (adverb+participle), (adjective+noun), (noun+(conjunction/preposition)+noun). Whenever a pattern appears in a sentence, the program checks for a collocation in Thesaurus and harvests its meaning.

---

[2] We understand a seme as a minimal bit of meaning.

[3] In the current article, we will not consider algorithms we used to assign a Wordnet definition to a target word. This issue will be addressed in further research.

[4] By collocations, we mean "expressions of multiple words which commonly co-occur" [Bird et al. 2009].

[5] Sections are not always immediate subdivisions of a Class. Some Sections are grouped in Divisions.

[6] The inside parentheses show that this part of the phrase may be missing; a slash stands for "or".

The algorithm collects Section numbers for every word and collocation and removes duplicates (in Thesaurus homonyms proper can be assigned to different subdivisions in the same or different sections), excluding stop words like "to", "a" etc.[7] Table 1 illustrates to what sections words in Ex. 1 belong.

**Table 1.** Semantic fields in the Banker joke

| Word | Section No., Section name in Thesaurus |
|---|---|
| I | — |
| use | 24, Volition In General <br> 30, Possessive Relations |
| to | — |
| be | 0, Existence <br> 19, Results Of Reasoning |
| a | — |
| banker | 31, Affections In General <br> 30, Possessive Relations |
| but | — |
| lose | 21, Nature Of Ideas Communicated <br> 26, Results Of Voluntary Action <br> 30, Possessive Relations <br> 19, Results Of Reasoning |
| interest | 30, Possessive Relations <br> 25, Antagonism <br> 24, Volition In General <br> 7, Causation <br> 31, Affections In General <br> 16, Precursory Conditions And Operations <br> 1, Relation |

Then the semantic vector of a pun is calculated. Every pun is a vector in a 34 dimensional space:

$$p_i = p_i(s_{1i}, s_{2i}, \ldots, s_{34i}) \quad (1)$$

The value of every element $s_{ki}$ equals the number of words in a pun, which belong to a Section $S_k$:

$$S_{ki} = \sum_{j=1}^{l_i} \{1 | w_{ji} \in S_k\}, \quad k = 1,2,\ldots,34, \quad i = 1,2,3\ldots \quad (2)$$

For example, the semantic vector of the Banker joke looks as follows:

$$p_{Banker} = \{1,1,0,0,0,0,0,1,0,0,0,0,0,0,0,1,0,0,2,0,1,0,0,2,1,1,0,0,0,4,2,0,0\}$$

To test the algorithm, we, first, collected 2,480 puns from different Internet resources and, second, built a corpus of 2,480 random sentences of length 5 to 25 words from different NLTK [Bird et al. 2009] corpora[8] plus several hundred aphorisms and proverbs from different Internet sites. We shuffled and split the sentences into two equal groups, the first two forming a training set and the other two a test set. The classification was conducted using different Scikit-learn [Pedregosa et al. 2011] algorithms. In all the tests, the Scikit-learn algorithm of SVM with the Radial Basis Function (RBF) kernel produced the highest average F-measure results ($\overline{f} = \frac{f_{puns} + f_{random}}{2}$). In addition, its results are smoother, comparing the difference between precision and recall (which leads to the highest F-measure scores) within the two classes (puns and random sentences), and between the classes (average scores).

---

[7] Stopwords are excluded from semantic analysis, but not from collocation extraction.

[8] Mainly Reuters, Web corpus and Gutenberg.

Table 2 illustrates results of different Scikit-learn algorithms, applied in classification of puns against two selections of random sentences: the first one (**Mixed styles**) is a mixture of Brown, Reuters and Web NLTK corpora, the second one (**Belles lettres**) contains sentences from Gutenberg (also NLTK), some proverbs and aphorisms. As the learning algorithms are widely used in NLP, we provide only their names. Their full description can be found in Scikit-learn documentation [Pedregosa et al. 2011]. The results given are a mean of five tests.

**Table 2.** Tests for pun recognition

|  | Precision |  | Recall |  | F-measure |  |
|---|---|---|---|---|---|---|
| **Method** | Pun | Not pun | Pun | Not pun | Pun | Not pun |
| **Mixed styles** | | | | | | |
| SVM with linear kernel | 0.68 | 0.66 | 0.63 | **0.71** | 0.66 | 0.68 |
| SVM with Radial Basis Function (RBF) kernel | **0.70** | **0.68** | **0.67** | **0.71** | **0.68** | **0.70** |
| **Belles lettres** | | | | | | |
| SVM with linear kernel | **0.75** | 0.69 | 0.65 | **0.78** | 0.69 | 0.73 |
| SVM with Radial Basis Function (RBF) kernel | 0.74 | **0.73** | **0.72** | 0.74 | **0.73** | **0.74** |
| Logistic Regression | 0.74 | 0.70 | 0.67 | 0.76 | 0.70 | 0.73 |

All the algorithms worked better in comparison of puns to literature, proverbs and aphorisms, the performance increasing by several percent. Moreover, within each class, SVM with the RBF kernel produced most of the highest results. The reason for this is most likely caused by the topicality issue: compared to random sentences many puns tackle similar issues, and even use recurring realias (for example, John Deere, appearing in 7 different puns). To see how big its influence is, we changed vectors, sorting numbers in them in a decreasing order, and retested the algorithms. First, the whole vector was sorted out (**First sorting** in Table 3); second, the initial vector was split into four parts of sizes 8, 8, 8, 10, and sorting was done within each part[9] (**Second sorting** in Table 3).

---

[9] The vector of the Banker joke now looks as follows: {1, 1, 1, 0, 0, 0, 0, 0 | 0, 0, 0, 0, 0, 0, 0, 0 | 2, 1, 1, 0, 0, 0, 0, 0 | 4, 2, 2, 1, 1, 0, 0, 0, 0, 0}.

Table 3. Tests for pun recognition: reduced topicality

| Method | Precision | | Recall | | F-measure | |
|---|---|---|---|---|---|---|
| | Pun | Not pun | Pun | Not pun | Pun | Not pun |
| First sorting | | | | | | |
| SVM with linear kernel | 0.57 | 0.56 | 0.56 | 0.57 | 0.56 | 0.57 |
| SVM with Radial Basis Function (RBF) kernel | **0.58** | **0.58** | **0.57** | **0.59** | **0.58** | **0.58** |
| Second sorting | | | | | | |
| SVM with linear kernel | **0.70** | 0.64 | 0.57 | **0.75** | 0.63 | **0.69** |
| SVM with Radial Basis Function (RBF) kernel | 0.68 | **0.66** | **0.64** | 0.70 | **0.66** | 0.68 |

After sorting, the difference between RBF and linear kernel becomes very low, but RBF is still (and inexplicably) more successful. The first sorting results in 58% success on average, when chance classification would produce a 50% result. The difference in 8% shows the purely structural potential of the algorithm, which, probably, rises from the curve of the semantic vector (differences among the most representative semantic fields, the "tail" of less representative fields, etc.). The partitioned sorting increases results by 10%, although the vector splits into four parts only. Splitting the vector into three parts results in 1% rise (not reflected in the table), which can be a feature of this particular dataset, or a more general trend, but this hypothesis requires more research.

The decrease in results shows that topicality is of much influence in pun recognition, although by definition a pun is not sense biased. This brings us to the idea that topicality is influential in puns as a *humorous* genre. Judging from the definition of a pun, as a means of expression, it can occur in any semantic context. However, puns, as a humorous genre, must inherit topical trends of humor. Some theories discuss existence of such trends. For example, R. Mihalcea and C. Strapparava write of "weak" human moments and targeting "professional communities that are often associated with amusing situations, such as lawyers, programmers, policemen like" in one-liners [Mihalcea et al. 2006: 139]. In [Mikhalkova 2009], we studied topical trends of physical and mental disorders, disorderly behavior, courtship, eating habits, and some other in comic TV-shows; etc.

## 3. Hitting the target word

We suggest that in a homographic pun the target word is a word that immediately belongs to two semantic fields; in a heterographic pun the target word belongs to at least one discovered semantic field and does not belong to the other. However, in reality, words in a sentence tend to belong to too many fields, and they create noise in the search. To reduce influence of noisy fields in the model, we included such a non-semantic feature as the tendency of the target word to occur closer to the end of a pun [Miller et al. 2015].

A-group ($W_A$) and B-group ($W_B$) are groups of words in a pun, which belong to the two semantic fields, sharing the target word. A-group attracts the maximum number of words in a pun:

$$S_{Ai} = \max_{k} S_{ki}, \quad k = 1,2,...,34 \qquad (3)$$

In the Banker joke $s_{Ai} = 4$, $A = 30$ (Possessive Relations); words that belong to this group are 'use', 'lose', 'banker', 'interest'.

B-group is the second largest group in a pun:

$$S_{Bi} = \max_{k}(S_{ki} \setminus S_{Ai}), \quad k = 1,2,...,34 \qquad (4)$$

In the Banker joke $s_{Bi} = 2$. There are three groups of words that have two words in them: $B_1 = 19$, Results Of Reasoning: "be", "lose"; $B_2 = 24$, Volition In General: "use", "interest"; $B_3 = 31$, Affections In General: "banker", "interest". Ideally there should be a group of about three words and collocations, describing a person's inner state ("used to be", "lose", "interest"), and two words ("lose", "interest") in $W_A$ are a target *phrase*. However, due to the shortage of data about collocations in dictionaries and limitations of the collocation extraction algorithm, $W_B$ divides into several smaller groups. Consequently, to find the target word, we appeal to other word features. In testing the system on homographic puns, we relied on polysemantic character of words. If in a joke, there are more than one value of $B$, $W_B$ candidates merge into one, with duplicates removed, and every word in $W_B$ becomes the target word candidate: ($c \in W_B$). In Ex. 1 $W_B$ is a list of "be", "lose", "use", "interest", "banker"; $B = \{19,24,31\}$.

Based on the definition of the target word in a homographic pun, words from $W_B$, that are also found in $W_A$, should have a privilege. Therefore, the first value ($v_\alpha$), each word gets, is the output of the Boolean function:

$$v_\alpha(c) = g(c, W_A, W_B) = \begin{cases} 2, & if\,(c \in W_A) \wedge (c \in W_B) \\ 1, & if\,(c \notin W_A) \wedge (c \in W_B) \end{cases} \qquad (5)$$

The second value ($v_\beta$) is the absolute frequency of a word in $W_{B_1} \cup W_{B_2} \cup W_{B_3}$ (the union of $B_1$, $B_2$, etc., including duplicates:

$$v_\beta(c) = f_c(W_{B_1} \cup W_{B_2} \cup W_{B_3})$$

Together values $v_\alpha$ and $v_\beta$ compose a group of sense criteria. In case of target word candidates, we multiply them and choose the word with the maximum rate:

$$z_1(W_B) = \{c \mid \max_{c}(v_\alpha \times v_\beta)\} \qquad (6)$$

The reasons for using plain multiplication in the objective function (6) lie in our treatment of puns properties. In the algorithm, they are maximization criteria: the more properties the sentence has and the more represented they are, the more likely the sentence is a pun. Grounded by maximization criteria, the word with the maximum rate is, therefore, the best candidate for the target word. In case of a tie, the algorithm picks up a random candidate.

Another way to locate the target word is to rely on its position in a pun $v_\gamma$: the closer it is to the end, the bigger this value is. If the word occurs several times, the algorithm counts the average of sums of position numbers. The output is again the word with the maximum value.

Values of the Banker joke are illustrated in Table 4.

**Table 4.** Values of the Banker joke

| Word form | $v_\alpha$ | $v_\beta$ | $z_1(W_B)$ | $v_\gamma$ |
|---|---|---|---|---|
| be | 1 | 1 | 1 | 4 |
| lose | 2 | 1 | 2 | 9 |
| use | 2 | 1 | 2 | 2 |
| interest | 2 | 2 | 4 | **10** |
| banker | 2 | 1 | 2 | 6 |

As for heterographic puns, the target word is missing in $W_B$ (the reader has to guess the word or phrase, homonymous to the target word). Accordingly, we rely on the completeness of the union of $W_A$ and $W_B$: among the candidates for $W_B$ (second largest groups) such groups are relevant, that form the longest list with $W_A$ (duplicates removed). In Ex. 2 (the Church joke) $W_A$ = {'go', 'gas', 'annual', 'barbecue', 'propane'}, and two groups form the largest union with it: $W_B$ = {'buy', 'proceeds'} + {'sacred', 'church'}. Every word in $W_A$ and $W_B$ can be the target word.

Due to sorting conditions, frequencies are of no value here; therefore, the method uses only the value of position in the sentence $v_\gamma$. The function output is:

$$z_2(W_A W_B) = \left\{ c \mid \max_c(v_\gamma) \right\} \qquad (7)$$

Values of the Church joke are illustrated in Table 5.

**Table 5.** Values of the Church joke

| Word form | $v_\gamma$ |
|---|---|
| propane | **18** |
| annual | 8 |
| gas | 5 |
| sacred | 15 |
| church | 3 |
| barbecue | 9 |
| go | 12 |
| proceeds | 11 |
| buy | 4 |

We tested the suggested algorithms on SemEval Gold data. Table 6 illustrates percentage of correct guesses within a pun (True Positive results).

SemEval organizers suggested their baselines for this task: selecting 1) a random word, 2) the last word in a pun, 3) the word with the biggest number of senses (the most polysemantic word) [Miller et al. 2017]. We also include their results in the table.

Table 6. Test results of target word analysis

|  | Homographic puns | Heterographic puns |
|---|---|---|
| Sense-based method, $z_1(W_B)$ | 0.2373 | — |
| Last word method, $v_\gamma$ | **0.5145** | 0.3879 |
| SemEval random | 0.0846 | 0.0839 |
| SemEval last word | 0.4704 | **0.5704** |
| SemEval polysemantic word | 0.1798 | 0.0110 |

Concerning homographic puns, the Last word method appears to be more effective, compared to SemEval last word, probably, due to the lack of filter for content words. At the same time, our Sense-based method is more effective than SemEval polysemantic word.

The Last word solution for heterographic puns turns out to be 18% less effective, than SemEval baseline (0.39 and 0.57, correspondingly). Testing heterographic puns with the algorithm for homographic puns brought even lower results. The reason for it, probably, lies in the method itself, that lacks the sense criterion about the target word present in one semantic group and absent in the other. This will be the only significant difference from the solution for homographic puns, beside a special treatment of $W_B$.

## 4. Results of SemEval-2017

Tables 7 and 8 reflect the top-scoring results of SemEval-2017, Task 7: Detection and Interpretation of English Puns, given in [Miller et al. 2017], and results of the own system PunFields (at competition and currently). Table 7 shows results for the class of puns. Table 8 shows Precision.

Table 7. SemEval pun classification

|  | Homographic puns | | | Heterographic puns | | |
|---|---|---|---|---|---|---|
|  | Precision | Recall | F-measure | Precision | Recall | F-measure |
| Duluth | 0.7832 | 0.8724 | 0.8254 | 0.7399 | 0.8662 | 0.7981 |
| Idiom Savant | — | — | — | **0.8704** | 0.8190 | **0.8439** |
| JU_CSE_NLP | 0.7251 | 0.9079 | 0.8063 | 0.7367 | **0.9402** | 0.8261 |
| N-Hance | 0.7553 | **0.9334** | **0.8350** | 0.7725 | 0.9300 | **0.8440** |
| PunFields | **0.8019** | 0.7785 | 0.7900 | 0.7585 | 0.6326 | 0.6898 |
| PunFields, current result[10] | 0.75 | 0.72 | 0.73 | 0.75 | 0.72 | 0.73 |

---

[10] Currently, we do not test homographic and heterographic puns separately.

Table 8. SemEval pun location

|  | Homographic puns | Heterographic puns |
|---|---|---|
| Idiom Savant | **0.6636** | 0.6845 |
| U-Waterloo | 0.6526 | **0.7973** |
| N-Hance | 0.4269 | 0.6592 |
| PunFields | 0.3279 | 0.3501 |
| PunFields, current result | 0.5145 | 0.3879 |

PunFields participated in SemEval-2017, Task 7 in a slightly different form. In pun classification (Paragraph 2), together with the collection of 2,480 puns, it used the Belles lettres corpus as a training set. In the present research the training set is twice smaller. Hence, the difference in results. The current result for pun location (Paragraph 3) is more valid, due to rethinking of sorting criteria and elimination of minor coding errors.

Generally, PunFields was most successful in pun classification, which can be due to advantages of supervised learning. Although there were other less successful systems, also using supervised learning algorithms.

SemEval winning systems in pun classification did not have much in common. Duluth used several WordNet customizations, some designed by its author T. Pedersen [Pedersen et al. 2009]. When these customizations disagree, the sentence is classified as a pun. IdiomSavant is a combination of different methods, including word2vec. JU_CSE_NLP is a supervised learning classifier, combining a hidden Markov model and a cyclic dependency network. N-Hance is a heuristic, making use of Pointwise Mutual Information, calculated for a list of word pairs[11]: the algorithm sorts out sentences, where the highest PMI is distinctively higher than its lower neighbor.

Concerning pun location, there were two systems that outperformed SemEval baseline by nearly 20%: Idiom Savant, described above, and UWaterloo. UWaterloo has 11 criteria to calculate the target word (word frequency, part-of-speech context, etc.), but again focuses on the second half of a pun. The system description papers have not been released so far, and it is hard to work out the main factor in the success of these two systems.

It is of interest that the simple approach, suggested by N-Hance, turned out to be so effective. Unlike other winners, it is not a supervised learning classifier or a combination of methods, some of which can be supple to tuning into a dataset. However, it was not as effective in pun location as in classification, and again the search was done among second elements of the pair with the highest PMI score (the end of the sentence criterion).

## Corollaries

We consider that the results of the present research allow us to state the following: the hypothesis about two semantic fields, underlying in every pun, is relatively true and objective; Roget's Thesaurus is a credible source in automatic semantic analysis; the semantic nature of puns (and other kinds of metaphorical language issues) can be subject to exact sciences. The suggested algorithm of pun detection and interpretation is fairly effective, but requires improvement. We tend to think that PunFields has advantageous prospects in customizing it to WordNet.

---

[11] PMI measure was calculated on the basis of a Wikipedia corpus.

The research also objectivizes some fundamental issues in understanding humor. One of them is topicality bind. There have been many suppositions and separately collected facts that humor is not universal, and that it thrives on some topics better than on other. Our pun classifier supports this trend.

In addition, we would like to stress the importance of phrases in creation of lexical ambiguity. Even in puns, where only one word is obviously ambiguous, its neighbors can have shades of other possible meanings. In the Banker joke, "lose" in collocation with "interest" can be antonym to "win, earn" in connection with "money, benefit", and to "get, gain" in connection with "curiosity".

Concerning location of the target word in a pun, competition results show that the structural "closer to the end" criterion is of great importance and is hard to beat even as the baseline. This issue has also been discussed in theories of humor: punchlines and target words do tend to occur at the end of an utterance.

SemEval competition included one more task: assigning a WordNet definition to the target word. This task appeared to be the most difficult, and very few systems beat the baseline results, which also leaves us grounds for further work.